\documentclass[runningheads]{llncs}
\usepackage[T1]{fontenc}
\usepackage{graphicx}
\usepackage{amsmath}
\usepackage{amssymb}
\usepackage{bm} 
\usepackage{bbm}
\usepackage{subcaption}

\usepackage{multirow}
\usepackage{pifont} 
\usepackage{hyperref}

\newcommand{\IT}[1]{``{\fontfamily{cmtt}\selectfont #1}''}

\usepackage{booktabs}
\usepackage{siunitx}
\usepackage{adjustbox}  
\usepackage{cleveref}

\newcommand{\cmark}{\ding{51}} 
\newcommand{\xmark}{\ding{55}} 

\crefname{figure}{Fig.}{Figs.}
\crefname{table}{Table}{Tables}
\crefname{section}{Sec.}{Secs.}
\crefname{equation}{Eq.}{Eqs.}

\begin{document}
\title{Hierarchical Co-Embedding of Font Shapes and Impression Tags}
%
%



\author{Yugo Kubota \and
Kaito Shiku \and
Seiichi Uchida}
\authorrunning{Y. Kubota et al.}

\institute{Kyushu University, Fukuoka, Japan \\
\email{yugo.kubota@human.ait.kyushu-u.ac.jp}}




\maketitle              
\begin{abstract}
Font shapes can evoke a wide range of impressions, but the correspondence between fonts and impression descriptions is not one-to-one: some impressions are broadly compatible with diverse styles, whereas others strongly constrain the set of plausible fonts. We refer to this graded constraint strength as \emph{style specificity}. In this paper, we propose a hyperbolic co-embedding framework that models font--impression correspondence through entailment rather than simple paired alignment. Font images and impression descriptions, represented as single tags or tag sets, are embedded in a shared hyperbolic space with two complementary entailment constraints: impression-to-font entailment and low-to-high style-specificity entailment among impressions. This formulation induces a radial structure in which low style-specificity impressions lie near the origin and high style-specificity impressions lie farther away, yielding an interpretable geometric measure of how strongly an impression constrains font style. Experiments on the MyFonts dataset demonstrate improved bidirectional retrieval over strong one-to-one baselines. In addition, traversal and tag-level analyses show that the learned space captures a coherent progression from ambiguous to more style-specific impressions and provides a meaningful, data-driven quantification of style specificity.

\end{abstract}



\section{Introduction}
\label{sec:intro}

\begin{figure}[t] 
    \centering
    \includegraphics[width=\linewidth]{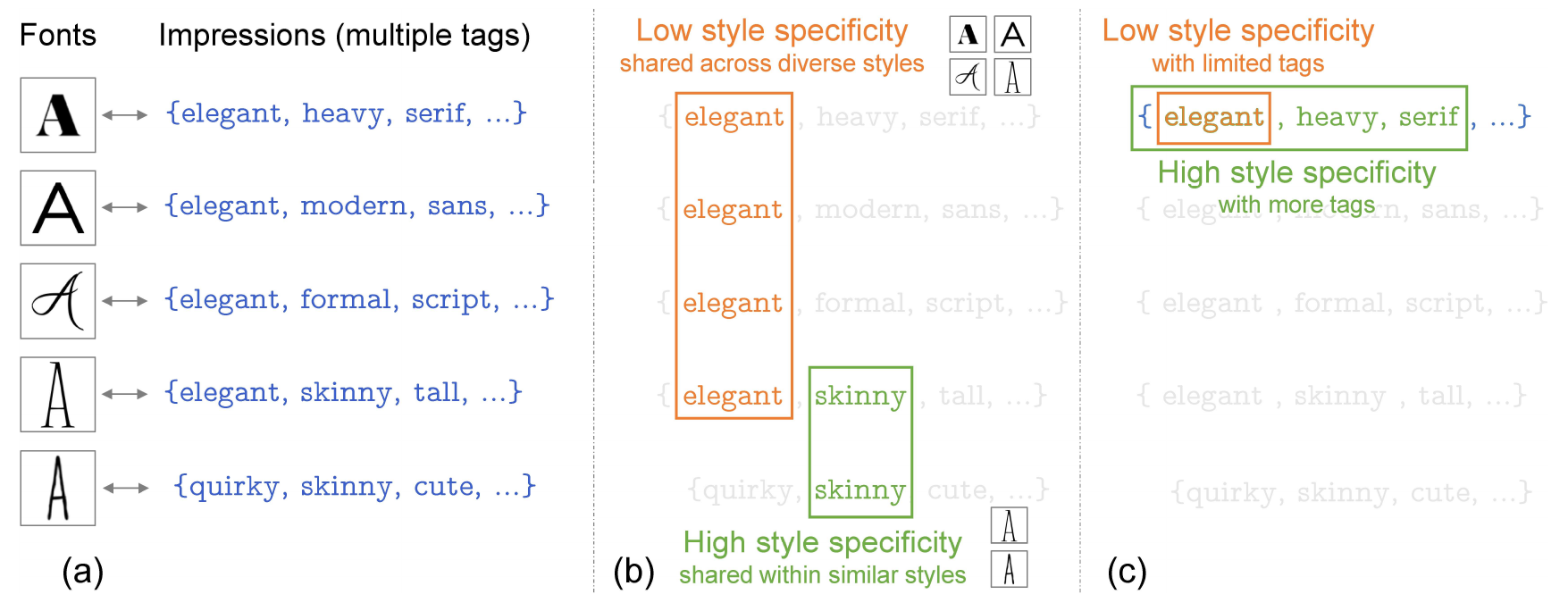}
    \caption{Fonts and their associated impression tags in the MyFonts dataset~\cite{MyFonts}. (a)~Fonts are annotated with sets of impression tags. (b) Single tags vary in style specificity, from broadly shared (low) to highly style-specific (high). (c) Adding complementary tags typically increases style specificity.}
    \label{fig:fonts}
\end{figure} 


Fonts are contour-defined shapes, yet they evoke a wide range of impressions. As shown in \cref{fig:fonts}(a), MyFonts~\cite{MyFonts}, an online font marketplace, allows users to freely annotate each font with tags. These tags span diverse concepts, ranging from categorical labels such as \IT{serif} and \IT{sans} to impression descriptors such as \IT{elegant} and \IT{cute}. In practice, a font’s perceived impression is typically expressed by a set of these tags rather than a single word. Throughout this paper, we refer to all these tags as \emph{impression tags.}
\par

As shown in \cref{fig:fonts}(b) and (c), we study how strongly impression descriptions, whether a single tag or a set of tags, constrain font style. We refer to this strength as \emph{style specificity}. In \cref{fig:fonts}(b), tags such as \IT{elegant} have low style specificity and are shared across diverse styles, whereas tags such as \IT{skinny} have high style specificity and concentrate within a much narrower style range. \cref{fig:fonts}(c) further suggests that style specificity depends on description granularity: a limited tag set often yields low specificity, while adding complementary tags typically makes the description more specific and increases style specificity. Importantly, style specificity is not explained by tag cardinality alone, since even single-tag descriptions can be either low- or high-specificity, as explained by \cref{fig:fonts}(b).
\par

Estimating the strength of font–impression correspondence is meaningful from both applied and cognitive-science perspectives. From an application standpoint, it helps designers understand what impressions a font can convey from its shape alone, enabling impression-aware font search, recommendation for a given textual context, and font generation conditioned on desired impressions. From a cognitive-science perspective, even relatively simple designs such as fonts can evoke diverse impressions from subtle shape differences. This suggests that humans acquire systematic associations between shape and impression, either through innate priors or learned experience. Quantifying these associations in a data-driven manner can contribute to the perceptual psychology of shape.
\par

However, it is not obvious how to quantify the strength of font–impression correspondence, and this issue has received limited attention in prior work. A common computational paradigm is to learn a shared embedding space for font shapes and impression text ({\em co-embedding}) from paired font–tag annotations, typically by optimizing a similarity-based alignment objective. For example, Impression-CLIP~\cite{ImpressionCLIP} focuses on aligning paired fonts and impressions, but does not explicitly distinguish between style-invariant and style-specific impressions. As a result, impressions with high and low style specificity can be treated similarly, making correspondence strength implicit at best. In contrast, cognitive-science approaches could, in principle, probe this strength via human judgments, but such studies do not scale to the large vocabularies of impression tags and the breadth of font styles encountered in real-world datasets.
\par

\begin{figure}[t] 
    \centering
    \includegraphics[width=0.97\linewidth]{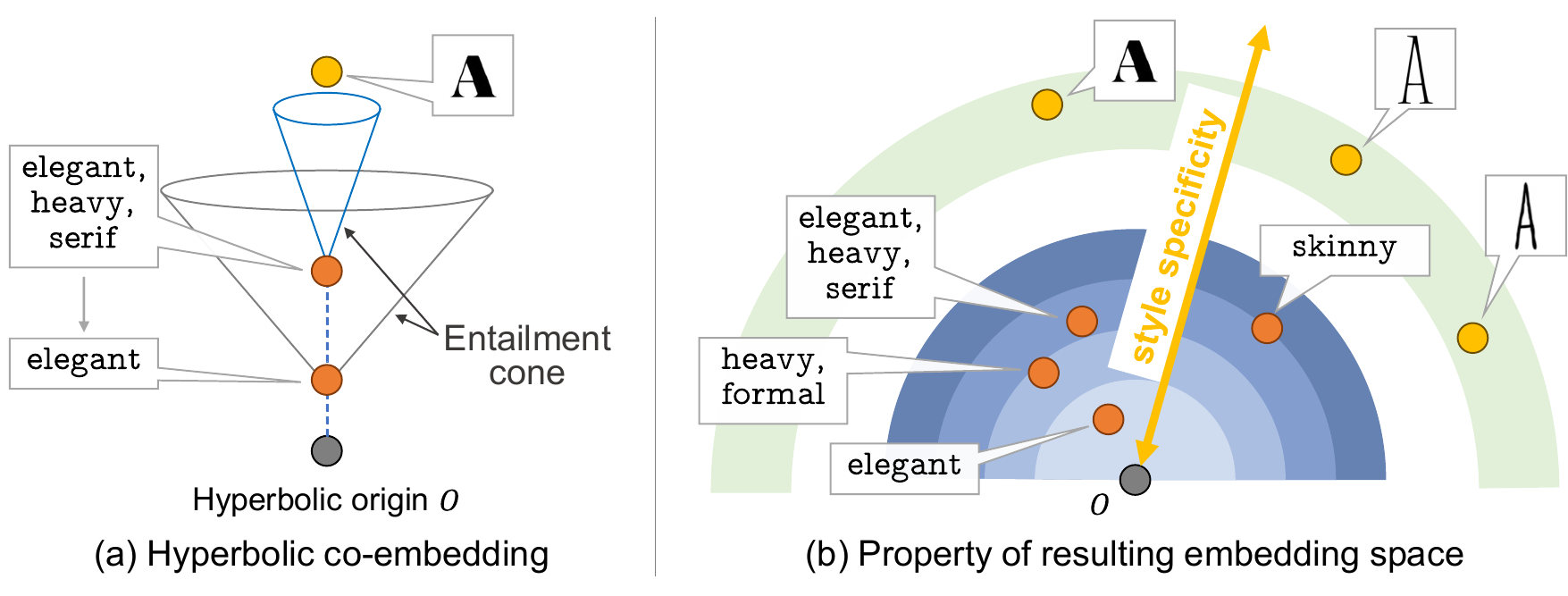}
    \caption{Hyperbolic co-embedding with entailment cones. (a)~Feature space where fonts and impression descriptions (tags or tag sets) are co-embedded, capturing impression-to-font entailment and style-specificity entailment among impressions. (b)~Radius encodes style specificity from low to high, as the result of hyperbolic co-embedding.}
    \label{fig:scheme1}
\end{figure}

In this work, we propose a {\em hyperbolic co-embedding} framework that automatically derives the strength of font–impression correspondence from large-scale data. As shown in \cref{fig:scheme1}(a), we jointly embed font images and impression descriptions, either single tags or tag sets, into a hyperbolic space~\cite{nickel2017poincare,nickel2018learning,sala2018representation} anchored at an origin. Similar to CLIP~\cite{CLIP}, strongly corresponding font images and impression descriptions are embedded close to each other. Unlike CLIP, our embedding assigns a semantic role to the radial coordinate. The distance to the origin encodes \emph{style specificity}: descriptions farther from the origin have higher style specificity, while those nearer the origin have lower style specificity. As illustrated in \cref{fig:scheme1}(b), font images lie at larger radii, and impression descriptions align along rays from the origin toward each font, progressing from low- to high-specificity.
\par

As illustrated in \cref{fig:scheme1}(a), our method defines \emph{entailment cones} in hyperbolic space to represent the region of font styles covered by each impression. This construction captures two types of entailment relationships: (i) \emph{impression-to-font entailment} and (ii) \emph{style-specificity entailment among impressions}, which links ambiguous descriptions (e.g., {\IT{elegant}}) to more specific ones (e.g., \{\IT{elegant}, \IT{heavy}, \IT{serif}\}). The former specifies which style region an impression can plausibly match. The latter encodes the ordering from low to high style specificity, where a more specific description is entailed by a more ambiguous one.
\par

\begin{figure}[t] 
    \centering
    \includegraphics[width=0.93\linewidth]{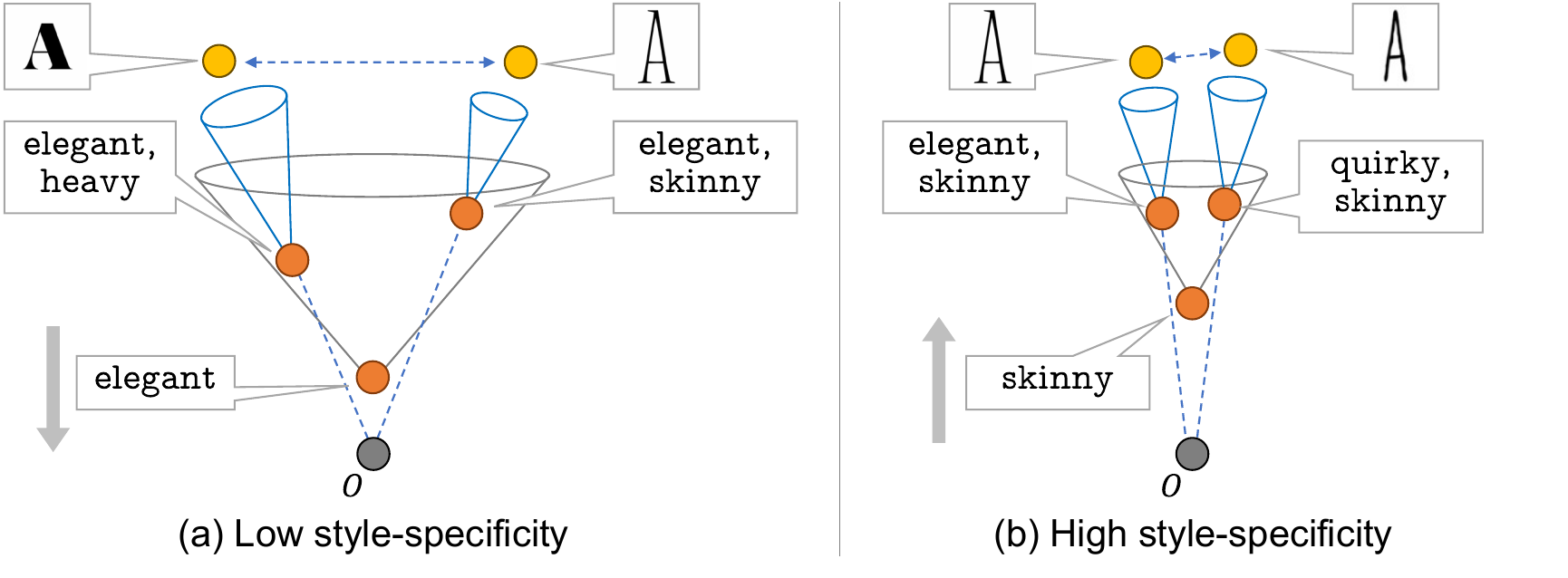}
    \caption{Cone aperture and style specificity. Cones widen near the origin for low-specificity tags (e.g., \IT{elegant}) and narrow at larger radii for high-specificity tags (e.g., \IT{skinny}), yielding broader vs. more constrained font-style coverage.}
    \label{fig:scheme2}
\end{figure}

The cone aperture decreases as an impression is embedded farther from the origin. Consequently, as shown in \cref{fig:scheme2}, broadly used tags with low style specificity such as \IT{elegant} appear near the origin, whereas highly style-specific tags such as \IT{skinny} lie at larger radii. With this geometry, the direction from the origin toward a font captures its impression profile, and the radial distance from the origin quantifies correspondence strength, namely style specificity.
\par

In our experiments on MyFonts dataset~\cite{MyFonts}, we show that the proposed hyperbolic co-embedding captures font–impression correspondence beyond one-to-one alignment. The learned space improves bidirectional retrieval and yields an interpretable notion of \emph{style specificity}, enabling an automatic, data-driven quantification of how strongly a tag or tag set constrains font style. Our main contributions are summarized as follows:

\begin{itemize}
\item We formulate font--impression correspondence in terms of two entailment relations: impression-to-font entailment and low-to-high style-specificity entailment among impression descriptions.
\item We propose a hyperbolic co-embedding framework with entailment cones that jointly embeds font images and impression descriptions, while assigning an interpretable role to radial distance as a measure of style specificity.
\item We demonstrate improved bidirectional retrieval over strong one-to-one baselines, and show that the learned geometry supports a data-driven quantification and analysis of how strongly impression tags constrain font style.
\end{itemize}
\par

\section{Related Work}
\subsection{Font Shapes and Subjective Impressions}
Psychology and HCI studies have repeatedly shown that subtle variations in typeface shape can systematically modulate the impressions people report. Classic work dating back to the early 20th century~\cite{davis1933determinants,franken1923study} and later studies~\cite{Morrison2006Educational,Shaikh2006Perception,Ying2010Typeface} provide evidence that particular fonts (or font attributes) tend to evoke characteristic affective or personality-like descriptors. 
These associations have been exploited in marketing and packaging to communicate brand and product images~\cite{henderson2004impression,Maisa2004typefaces}, and in text-based communication to convey paralinguistic cues such as emotion or speaker intent~\cite{TypefaceEmotion,Emotype,CHUJO20242Exploring}. 
However, most findings are derived from small-scale user studies with limited font collections and impression vocabularies, leaving open how to quantify the strength of font--impression correspondence across diverse styles and large, open-ended sets of descriptors.
\par

\subsection{Fonts and Their Impressions in Machine Learning}

Font--impression associations have been studied not only in cognitive science but also in machine learning. O’Donovan et al.~\cite{o2014exploratory} constructed a dataset of fonts paired with impression scores and introduced an impression-driven font search interface. Chen et al.~\cite{MyFonts} later collected the larger-scale MyFonts dataset of fonts annotated with user-provided impression tags and proposed an impression-based retrieval model. 
More recently, these datasets have enabled impression-conditioned font retrieval and generation~\cite{matsuda2021impressions2font,matsuda2022font,Izumi2024CLIPFontDraw,GRIF-DM,kubota2025embedding,tatsukawa2024FontCLIP}.
\par
Several machine-learning methods learn a shared latent space for fonts and impression descriptors, enabling both applications and analysis of the correspondence between fonts and impressions.
Early joint-embedding work includes~\cite{choi2019assist,kulahcioglu2020fonts}, while more recent co-embedding methods use VAEs or contrastive learning to better model these relations~\cite{CrossAE,ImpressionCLIP}.
A key difference is that prior approaches largely treat correspondence as paired alignment between a font and an impression description, whereas we model it via entailment to capture the inherent one-to-many setting, where a single impression description can be compatible with multiple fonts.

\subsection{Hyperbolic Space and Its Application for Computer Vision}
Hyperbolic space is a complete, simply connected Riemannian manifold with constant negative sectional curvature~\cite{nickel2017poincare,nickel2018learning,sala2018representation}. Unlike Euclidean space, its volume grows exponentially with radius, making it particularly effective for representing tree-like conceptual hierarchies.
\par

Hyperbolic geometry has recently been adopted to encode hierarchies in a range of computer vision problems, including text-image alignment~\cite{desai2023hyperbolic,pal2024compositional}, image-object relation modeling~\cite{ge2023hyperbolic}, video frame retrieval~\cite{li2025enhancing}, anomaly detection~\cite{li2024hyperbolic}, and medical image analysis~\cite{gonzalez2025hyperbolic}. The most relevant to our setting is MERU~\cite{desai2023hyperbolic}, which targets the semantic hierarchy underlying image-text correspondence. MERU combines hyperbolic contrastive learning with a hierarchy-aware entailment loss that captures inclusion relations between text and images, yielding improved hierarchical consistency and stronger retrieval performance than Euclidean CLIP models~\cite{CLIP}.
\par

To the best of our knowledge, this work is the first to introduce hyperbolic representation learning for co-embedding fonts and impressions. Importantly, impression descriptions vary in granularity: some tags are highly style-specific and strongly constrain font shape, whereas others are more ambiguous and compatible with diverse styles. Hyperbolic geometry provides a natural inductive bias for such heterogeneous, abstraction-to-specific relationships, motivating our entailment-based, hierarchy-aware embedding framework.
\par

\section{Preliminaries on Hyperbolic Space}
\label{sec:preliminary}
To model the style specificity of impressions, we embed fonts and impression tags in hyperbolic space. As shown in \cref{fig:scheme2}, entailment cones induce a radial ordering that places low style-specificity impressions near the origin and high style-specificity impressions farther away. This section formalizes the hyperbolic geometry underlying our method.

Among equivalent models of hyperbolic space~\cite{nickel2017poincare,nickel2018learning,sala2018representation}, we use the Lorentz model~\cite{nickel2018learning} for stability.
The $d$-dimensional hyperbolic space of curvature $-c$ ($c>0$) is the upper sheet of a two-sheeted hyperboloid in $\mathbb{R}^{d+1}$ with the Lorentzian inner product:
\begin{equation}
\mathbb{L}^d
=
\left\{
\bm{x}\in\mathbb{R}^{d+1}
:\;
\langle \bm{x},\bm{x}\rangle_{\mathbb{L}}=-\frac{1}{c},
\;\;
x_{\mathrm{time}}>0
\right\}.
\end{equation}

Here, $\bm{x}\in\mathbb{R}^{d+1}$ denotes a font or impression embedding. We write $\bm{x}=[\bm{x}_{\mathrm{space}},x_{\mathrm{time}}]$, where $x_{\mathrm{time}}$ is the coordinate on the hyperboloid's symmetry axis and $\bm{x}_{\mathrm{space}}$ contains the others. The Lorentzian inner product for any $\bm{x},\bm{y}\in\mathbb{R}^{d+1}$ is
\begin{equation}
\langle \bm{x},\bm{y}\rangle_{\mathbb{L}}
=
\langle \bm{x}_{\mathrm{space}},\bm{y}_{\mathrm{space}}\rangle
-
x_{\mathrm{time}}\,y_{\mathrm{itme}}.
\end{equation}

In the Lorentz model, the geodesic distance is given by
\begin{equation}
d_{\mathbb{L}}(\bm{x},\bm{y})
=
\frac{1}{\sqrt{c}}\,
\cosh^{-1}\!\bigl(-c\,\langle \bm{x},\bm{y}\rangle_{\mathbb{L}}\bigr),
\qquad
\bm{x},\bm{y}\in\mathbb{L}^d.
\label{eq:distance}
\end{equation}

We fix the base point (the ``origin'') as
$\bm{o}=[\bm{0},\,1/\sqrt{c}]\in\mathbb{L}^d$.
The tangent space at $\bm{o}$ is
\begin{equation}
T_{\bm{o}}\mathbb{L}^d
=
\left\{
\bm{v}\in\mathbb{R}^{d+1}
:\;
\langle \bm{v},\bm{o}\rangle_{\mathbb{L}}=0
\right\},
\end{equation}
equipped with the induced (Riemannian) norm $\|\bm{v}\|_{\mathbb{L}}=\sqrt{\langle \bm{v},\bm{v}\rangle_{\mathbb{L}}}$ for $\bm{v}\in T_{\bm{o}}\mathbb{L}^d$.
The exponential map $\exp^{c}_{\bm{o}}:T_{\bm{o}}\mathbb{L}^d\to\mathbb{L}^d$ is
\begin{equation}
\exp^{c}_{\bm{o}}(\bm{v})
=
\cosh\!\left(\sqrt{c}\,\|\bm{v}\|_{\mathbb{L}}\right)\bm{o}
+
\frac{
\sinh\!\left(\sqrt{c}\,\|\bm{v}\|_{\mathbb{L}}\right)
}{
\sqrt{c}\,\|\bm{v}\|_{\mathbb{L}}
}
\,\bm{v}.
\label{eq:expmapping}
\end{equation}
\par

\section{Hierarchical Co-Embedding of Font Shapes and Impression Tags in Hyperbolic Space}
\label{sec:method}

\subsection{Overview}

\begin{figure}[t] 
    \centering
    \includegraphics[width=\linewidth]{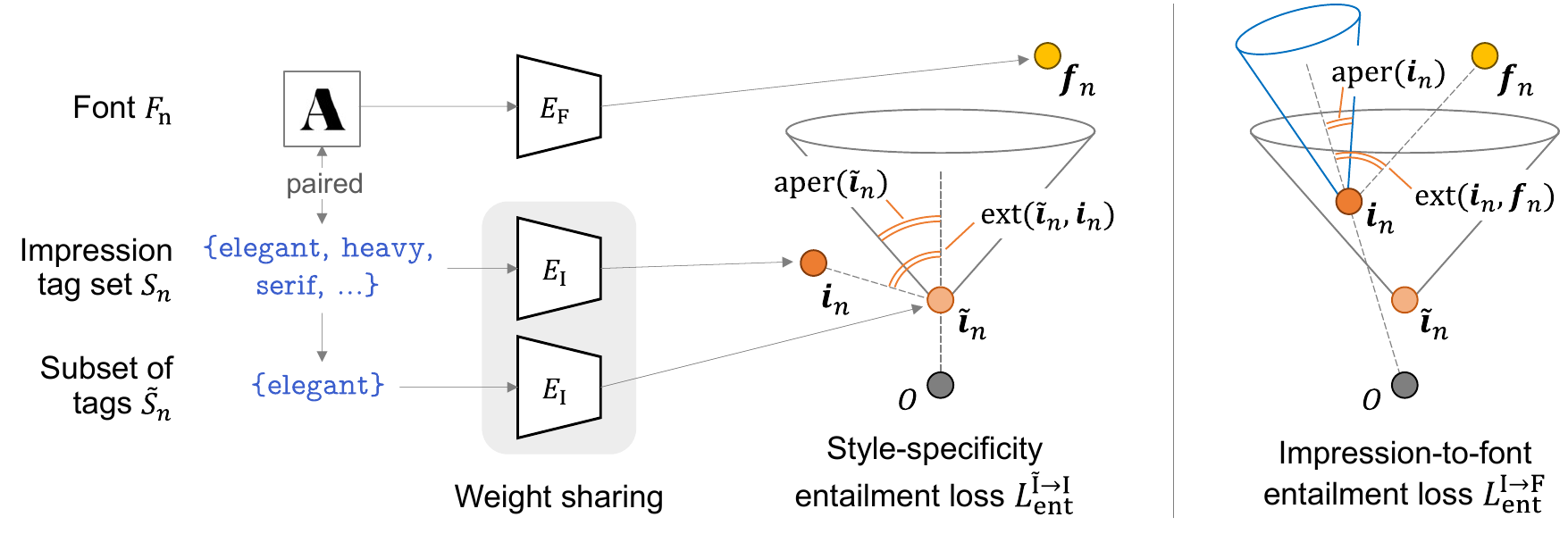}
    \caption{Overview. We embed a font $F_n$, its impression tag set $S_n$, and a subset $\tilde{S}_n$ in a shared hyperbolic space. Entailment cones place the lower style-specificity embedding of $\tilde{S}_n$ nearer the origin and the higher style-specificity embedding of $S_n$ farther away, enforcing $\tilde{\bm{i}}_n\!\to\!\bm{i}_n$. The cones also impose impression-to-font entailment $\bm{i}_n\!\to\!\bm{f}_n$ through $\mathrm{aper}(\cdot)$ and $\mathrm{ext}(\cdot,\cdot)$.}
    \label{fig:overview}
\end{figure} 
\cref{fig:overview} shows an overview of our approach. Our objective is to encode graded style-specificity as an entailment hierarchy, capturing both (i) entailment between impression descriptions and font shapes and (ii) low-to-high style-specificity entailment among impressions. To this end, we co-embed fonts and impressions in a shared hyperbolic space, whose geometry naturally supports continuous hierarchical organization. The resulting representation enables data-driven quantification of style specificity in the correspondence between impressions and font styles, providing a geometric basis for style consistency.
\par
%

To encode these entailment-based hierarchical relations in a continuous space, our learning objective combines two complementary losses. First, a hyperbolic contrastive loss aligns fonts and impressions across modalities in the shared space $\mathbb{L}^d$. Second, an entailment loss enforces the desired entailment relations on the embeddings, capturing both impression-to-font entailment and low-to-high style-specificity entailment among impressions. Together, these losses yield a shared space that is both cross-modally aligned and hierarchically organized.
\par

\subsection{Embedding Fonts and Impressions in Hyperbolic Space}
\label{sub-sec:embedding}
In this section, we describe how we embed font images and impression tag sets into hyperbolic space.
Our training data consist of pairs $\{(F_n,\mathcal{S}_n)\}_{n=1}^{N}$, where each font $F_n$ comprises 26 uppercase letter images and each $S_n=\{t_{n,k}\}_{k=1}^{K_n}$ is a set of impression tags drawn from a fixed vocabulary.
We extract feature representations using a font image encoder $E_{\mathrm{F}}$ and an impression tag encoder $E_{\mathrm{I}}$. Following \cref{sec:preliminary}, we regard the encoder outputs as tangent vectors at the origin $\bm{o}$ and map them onto the Lorentz model via the exponential map in \cref{eq:expmapping}:
\begin{equation}
\label{eq:expmap_font_imp}
\bm{f}_n = \exp_{\bm{o}}^{c}\!\left(E_{\mathrm{F}}(F_n)\right), \qquad 
\bm{i}_n = \exp_{\bm{o}}^{c}\!\left(E_{\mathrm{I}}(\mathcal{S}_n)\right).
\end{equation}
\par

To explicitly model the low-to-high style-specificity structure among impressions, we additionally construct a subset of each tag set, denoted $\tilde{\mathcal{S}}_n$, to serve as a lower-specificity impression description. We encode $\tilde{\mathcal{S}}_n$ in the same manner as $\mathcal{S}_n$, yielding the corresponding embedding $\tilde{\bm{i}}_n$.
\par

\subsection{Hyperbolic Co-Embedding of Fonts and Impressions via Contrastive Learning}
To obtain co-embedded representations of fonts and impressions in hyperbolic space, we first apply contrastive learning using the negative Lorentzian distance in \cref{eq:distance} as the similarity measure. The objective is to pull matched font--impression pairs closer in the shared space while pushing mismatched pairs farther apart.
\par

Given a mini-batch of paired embeddings $\{(\bm{f}_n,\bm{i}_n)\}_{n=1}^{B}$, we employ a bidirectional contrastive objective.
The impression-to-font loss is
\begin{equation}
\label{eq:cont_if_nomask}
\mathcal{L}_{\mathrm{cont}}^{\mathrm{I\rightarrow F}}
=-
\sum_{n=1}^{B}\log
\frac{\exp\!\left(-d_{\mathbb{L}}(\bm{i}_n,\bm{f}_n)/\tau\right)}
{\sum_{m=1}^{B}\exp\!\left(-d_{\mathbb{L}}(\bm{i}_n,\bm{f}_m)/\tau\right)},
\end{equation}
and the font-to-impression loss $\mathcal{L}_{\mathrm{cont}}^{\mathrm{F\rightarrow I}}$ is defined analogously by swapping the roles of fonts and impressions.
Here, $\tau>0$ is a learnable temperature parameter that controls the sharpness of the softmax distribution.
\par

Furthermore, to learn the correspondence between fonts and impressions with lower style specificity, we additionally introduce an alignment loss from the impression-tag subset to fonts, denoted by $\mathcal{L}_{\mathrm{cont}}^{\mathrm{\tilde I\rightarrow F}}$. This loss is defined analogously to the impression-to-font alignment loss $\mathcal{L}_{\mathrm{cont}}^{\mathrm{I\rightarrow F}}$, except that it uses embeddings computed from the subset impression tag set $\tilde{\mathcal{S}}_n$. Combining the above three losses, our overall contrastive objective is
\begin{equation}
L_{\mathrm{cont}}
=
\tfrac{1}{4}\mathcal{L}_{\mathrm{cont}}^{\mathrm{I\rightarrow F}}
+
\tfrac{1}{4}\mathcal{L}_{\mathrm{cont}}^{\mathrm{\tilde I\rightarrow F}}
+
\tfrac{1}{2}\mathcal{L}_{\mathrm{cont}}^{\mathrm{F\rightarrow I}}.
\end{equation}
\par


\subsection{Hierarchical Modeling of Impressions and Fonts via Entailment Learning}
\label{sub-sec:loss_ent}
%
To obtain a shared hyperbolic space in which low style-specificity impressions are placed closer to the origin $o$ and highly style-specific impressions are placed farther away, we leverage entailment cones. Specifically, we model two types of entailment: (i) impression-to-font entailment, which captures the relation between impressions and fonts, and (ii) entailment among impressions, which captures the ordering from lower to higher style specificity.
\par

We first review the notion of an entailment cone. Given an arbitrary point $\bm{x}\in\mathbb{L}^d$ in hyperbolic space, the cone associated with $\bm{x}$ is determined by its aperture $\mathrm{aper}(\bm{x})$, defined as
\begin{equation}
\label{eq:aper}
\mathrm{aper}\!\left(\bm{x}\right)
=
\sin^{-1}\!\left(\frac{2K}{\sqrt{c}\,\|\bm{x}_{\mathrm{space}}\|}\right),
\end{equation}
where $K$ controls the angular extent of the cone. Following MERU~\cite{desai2023hyperbolic}, we set $K=0.1$. This definition yields wider cones near the origin and narrower cones farther away, consistent with our objective that low style-specificity impressions admit broader entailments than highly style-specific ones.
\par

Given the entailment cone induced by $\bm{x}$, we determine whether an arbitrary point $\bm{y}$ lies inside it. If $\bm{y}$ falls outside the cone, we penalize the violation to enforce the desired entailment relation. To this end, we compute the exterior angle $\mathrm{ext}(\bm{x},\bm{y})=\pi-\angle o\bm{x}\bm{y}$ as
\begin{equation}
\label{eq:ext}
\mathrm{ext}(\bm{x},\bm{y})
=
\cos^{-1}\!\left(
\frac{
y_{\mathrm{time}} + x_{\mathrm{time}}\, c\, \langle \bm{x}, \bm{y} \rangle_{\mathbb{L}}
}{
\|\bm{x}_{\mathrm{space}}\|\,
\sqrt{\left(c\,\langle \bm{x}, \bm{y} \rangle_{\mathbb{L}}\right)^2 - 1}
}
\right),
\end{equation}
and penalize cone violations using the hinge loss
\begin{equation}
\label{eq:ent}
L_{\mathrm{ent}}(\bm{x},\bm{y})
=
\max\!\bigl(0,\,
\mathrm{ext}(\bm{x}, \bm{y})-\mathrm{aper}(\bm{x})
\bigr).
\end{equation}

For a mini-batch of paired embeddings $\{(\bm{f}_n,\bm{i}_n)\}_{n=1}^{B}$, we define the following entailment losses using the entailment-cone formulation above. Specifically, we enforce (i) impression-to-font entailment and (ii) low-to-high style-specificity entailment among impressions, defined as
\begin{equation}
\label{eq:ent_ii}
L_{\mathrm{ent}}^{\mathrm{I}\rightarrow\mathrm{F}}
= \sum_{n=1}^{B} L_{\mathrm{ent}}(\bm{i}_n,\bm{f}_n),
\quad
L_{\mathrm{ent}}^{\tilde{\mathrm{I}}\rightarrow\mathrm{I}}
= \sum_{n=1}^{B} L_{\mathrm{ent}}(\tilde{\bm{i}}_n,\bm{i}_n).
\end{equation}

Overall, our proposed method jointly optimizes the contrastive objective for bidirectional font--impression alignment and two entailment losses that impose entailment-aware constraints in the shared hyperbolic space. The total objective is
\begin{equation}
\label{eq:loss_total}
L_{\mathrm{total}}
=
L_{\mathrm{cont}}
+
\lambda_{1}\, L_{\mathrm{ent}}^{\mathrm{I}\rightarrow\mathrm{F}}
+
\lambda_{2}\, L_{\mathrm{ent}}^{\tilde{\mathrm{I}}\rightarrow\mathrm{I}},
\end{equation}
where $\lambda_{1},\lambda_{2}>0$ control the contributions of the two entailment terms. We set $\lambda_{1}=\lambda_{2}=0.1$ in all experiments.
\par


\section{Experimental Settings}


\subsection{MyFonts Dataset~\cite{MyFonts}}
We use the MyFonts dataset collected by Chen \textit{et al.}~\cite{MyFonts}, which contains 18{,}815 fonts annotated with open-vocabulary impression tags from \href{https://www.myfonts.com/}{MyFonts.com}, totaling 1{,}824 unique tags. We remove dingbat fonts (i.e., fonts consisting of pictorial symbols and lacking letter glyphs) and split the remaining 16{,}791 fonts into train/val/test sets with 13{,}461, 1{,}667, and 1{,}663 fonts, respectively. We restrict the impression-tag vocabulary to tags that appear at least 50 times in the training split, yielding 631 tags used throughout our experiments.
\par

\subsection{Implementation Details}
We use a ResNet-18~\cite{he2016deep} as the font image encoder $E_{\mathrm{F}}$ and a Transformer~\cite{vaswani2017attention} as the impression tag encoder $E_{\mathrm{I}}$.
For each font $F_n$, we render the 26 uppercase glyphs as $32{\times}32$ images, stack them, and feed the resulting tensor into the ResNet to obtain the font embedding $\bm{f}_n$.
For impressions, we first encode each tag with a frozen CLIP text encoder~\cite{CLIP} using the prompt ``The impression is \{tag\}.'' to obtain tag features; we then input these features, together with a \texttt{[CLS]} token, into the Transformer and use the \texttt{[CLS]} output as the impression embedding $\bm{i}_n$.
We optimize all trainable parameters with AdamW using a learning rate of $10^{-5}$ and a batch size of 32.
Following MERU~\cite{desai2023hyperbolic} for numerical stability, we additionally apply learnable scalar scaling to the outputs of both encoders.
\par

\subsection{Baselines}
\label{sub-sec:baselines}
We use two baselines, Impression-CLIP~\cite{ImpressionCLIP} and Cross-AE~\cite{CrossAE}. Impression-CLIP is a CLIP-style co-embedding method that learns font--impression alignment on the hypersphere via a bidirectional contrastive objective~\cite{CLIP,ImpressionCLIP,tatsukawa2024FontCLIP}. Cross-AE also uses the same encoders, but additionally introduces decoders and couples font and impression representations through within-modality reconstruction using paired font--impression data. Unlike our approach, these baselines do not explicitly model entailment between fonts and impressions; instead, they primarily treat the correspondence as one-to-one alignment. \par
For a fair comparison, we update Impression-CLIP and Cross-AE from their original implementations. Specifically, we reimplement both baselines using the same font and impression encoders as ours and train them on the identical train/val/test split. Hereafter, we call the updated versions, Impression-CLIP+ and Cross-AE+, respectively.
\par


\subsection{Evaluation Metrics}

In \cref{sec:quant}, we quantify how well a model captures font--impression correspondence by testing whether paired fonts and impression tag sets are placed close to each other in a shared latent space. If the space faithfully encodes this correspondence, a query from one modality should rank its corresponding items from the other modality ahead of non-corresponding ones. We therefore evaluate bidirectional retrieval on the test split in both directions, impression-to-font and font-to-impression, as a direct proxy for correspondence quality.
\par

We report two standard ranking metrics, mean Average Precision (mAP) and normalized Discounted Cumulative Gain (nDCG)~\cite{jarvelin2002cumulated}. For mAP, we evaluate impression-to-font retrieval under two query protocols: (i) \textbf{mAP-single}, where each of the 631 vocabulary tags is used as a single-tag query, and (ii) \textbf{mAP-multi}, where queries are tag sets. For \textbf{mAP-multi}, we generate 7{,}116 unique queries by sampling five subsets of size two to five tags from each test tag set and removing duplicates. In both protocols, a retrieved font is considered relevant if its tag set contains \emph{all} query tags.
\par


For nDCG, we evaluate top-100 ranking quality (nDCG@100) in both retrieval directions on the test split. Specifically, we use each test impression tag set as a query for impression-to-font retrieval, and we use each test font as a query for font-to-impression retrieval. For each query and candidate, we assign a graded relevance score based on tag-set agreement (precision, recall, or F1) between the query-side tag set and the candidate-side tag set, and then compute nDCG@100. We finally average nDCG@100 across queries.
\par

\section{Experimental Results}
\subsection{Quantitative Evaluation of Bidirectional Retrieval}
\label{sec:quant}

\begin{table}[t] 
\centering
\footnotesize
\renewcommand{\arraystretch}{1.12}
\setlength{\tabcolsep}{3.8pt}
\caption{Quantitative results on bidirectional retrieval between fonts and impressions. Top: baselines. Bottom: ours and ablations. Higher is better for all metrics. The best and second-best results are highlighted in boldface and underlined, respectively.}
\begin{adjustbox}{max width=\linewidth}
\begin{tabular}{
@{}l c c c S S S S S S S S@{}}
\toprule
\multirow{2}{*}{Method}
& \multirow{2}{*}{$L_{\mathrm{cont}}^{\mathrm{\tilde I\to F}}$}
& \multirow{2}{*}{$L_{\mathrm{ent}}^{\mathrm{I\to F}}$}
& \multirow{2}{*}{$L_{\mathrm{ent}}^{\mathrm{\tilde I\to I}}$}
& \multicolumn{2}{c}{$\mathrm{mAP}^{\mathrm{I\to F}}$}
& \multicolumn{3}{c}{$\mathrm{nDCG@100}^{\mathrm{I\to F}}$}
& \multicolumn{3}{c}{$\mathrm{nDCG@100}^{\mathrm{F\to I}}$} \\
\cmidrule(lr){5-6}\cmidrule(lr){7-9}\cmidrule(lr){10-12}
& & & & {Single} & {Multi} & {Re.} & {Prec.} & {F1} & {Re.} & {Prec.} & {F1} \\
\midrule
Impression-CLIP+$^\star$
&        &        &        & 0.042 & 0.031 & \textbf{0.321} & 0.381 & 0.402 & \underline{0.315} & 0.369 & 0.400 \\
Cross-AE+$^{\star\star}$
&        &        &        & 0.039 & 0.019 & 0.216 & 0.253 & 0.269 & \textbf{0.341} & 0.404 & \textbf{0.426} \\

\midrule
\addlinespace[2pt]

\multirow{4}{*}{Ours}
& \xmark & \xmark & \xmark & 0.042 & 0.031 & 0.307 & 0.365 & 0.390 & 0.300 & 0.353 & 0.387 \\
& \cmark & \xmark & \xmark & 0.057 & 0.040 & 0.313 & 0.369 & 0.394 & 0.287 & 0.394 & 0.405 \\
& \cmark & \cmark & \xmark & \underline{0.078} & \underline{0.048} & 0.315 & \underline{0.389} & \underline{0.405} & 0.291 & \textbf{0.412} & 0.405 \\
& \cmark & \cmark & \cmark & \textbf{0.084} & \textbf{0.052} & \underline{0.320} & \textbf{0.400} & \textbf{0.414} & 0.306 & \underline{0.407} & \underline{0.408} \\

\bottomrule
\end{tabular}
\end{adjustbox}
\vspace{-4mm}
\flushleft{\scriptsize $\star$ Impression-CLIP+ is an updated version of Impression-CLIP\cite{ImpressionCLIP} for a fair comparison.}\\
\vspace{-4mm}
\flushleft{\scriptsize $\star\star$ Cross-AE+ is an updated version of Cross-AE\cite{CrossAE} for a fair comparison.}
\label{tab:ablation_results}
\end{table} 

\cref{tab:ablation_results} reports bidirectional retrieval results between fonts and impression tag sets. Our method outperforms the updated baselines on most metrics, with notably higher mAP for impression-to-font retrieval, indicating more reliable font search from sparse impression queries. While Impression-CLIP+ and Cross-AE+ mainly learn one-to-one alignment, our entailment-aware hyperbolic embedding with entailment cones better handles varying description granularity, yielding one-to-many correspondences in which low style-specificity impressions match diverse fonts and more style-specific impressions produce sharper retrieval.
\par

The bottom part of \cref{tab:ablation_results} presents an ablation of our method. Performance improves monotonically as we add loss terms. The contrastive-only variant in hyperbolic space provides a weak baseline, whereas incorporating $L_{\mathrm{cont}}^{\mathrm{\tilde I\to F}}$ improves mAP by better aligning fonts with ambiguous, low style-specificity queries. Adding the font--impression entailment term $L_{\mathrm{ent}}^{\mathrm{I\to F}}$ yields a further substantial gain, and the impression entailment term $L_{\mathrm{ent}}^{\mathrm{\tilde I\to I}}$ provides an additional improvement, consistent with the model capturing granularity differences among impression descriptions as entailment-aware structure. Overall, these components are complementary and promote entailment-based hierarchical learning.
\par

\subsection{How Are the Impression Tags Embedded? (1) Distribution along Hyperbolic Space}
\begin{figure}[t] 
    \centering
    \begin{subfigure}{0.48\linewidth}
        \includegraphics[width=\linewidth]{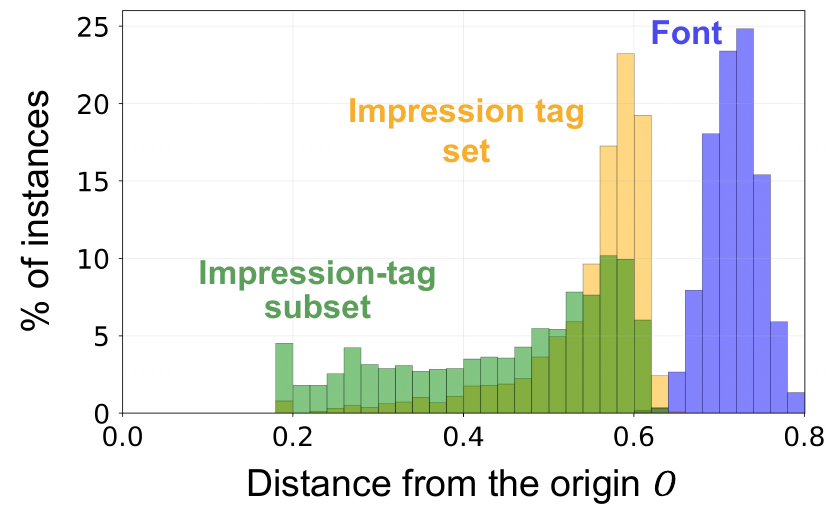}
        \caption{Proposed method}
        \label{fig:dist_from_root_proposed}
    \end{subfigure}
    \hfill
    \begin{subfigure}{0.48\linewidth}
        \includegraphics[width=\linewidth]{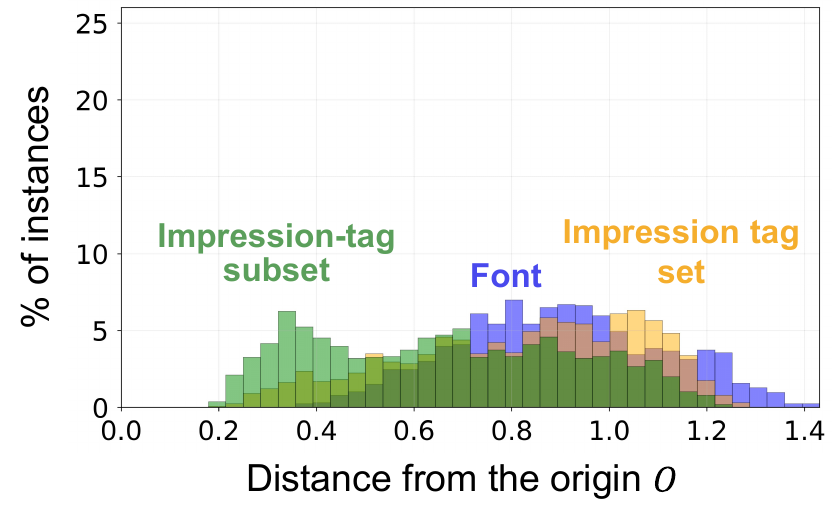}
        \caption{Impression-CLIP+}
        \label{fig:dist_from_root_CLIP}
    \end{subfigure}
    \caption{Histograms of distances from the origin $o$ for fonts, impression-tag sets, and impression-tag subsets on the test split. Our method shows a clear radial ordering, whereas Impression-CLIP+ exhibits substantial overlap.}
    \label{fig:dist_from_root}
\end{figure} 

To verify that our entailment-driven learning in hyperbolic space induces the intended hierarchy, in contrast to Impression-CLIP+ trained in Euclidean space, we analyze distance-to-origin distributions. Specifically, \cref{fig:dist_from_root} plots, on the test split, the distributions of distances from the origin $o$ for font embeddings, impression-tag-set embeddings, and embeddings of impression-tag subsets (i.e., more ambiguous descriptions induced by using fewer tags). For Impression-CLIP+, following MERU~\cite{desai2023hyperbolic}, we use as the reference point the normalized mean embedding computed over all samples. If the hierarchy is properly captured, fonts should concentrate at larger distances than impressions, and, among impressions, impression-tag subsets should lie closest to the origin.

These results show that Impression-CLIP+ yields substantial overlap among the three distance distributions, indicating limited hierarchical separation.
In contrast, our method exhibits a clear radial ordering from the origin: impression-tag subsets are closest, impression-tag sets lie farther, and fonts are farthest.
This consistent ordering supports that our hyperbolic embedding captures the intended entailment-driven structure.

\par

\subsection{How Are the Impression Tags Embedded? (2) Traversal along the Trajectory from the Origin $o$ to a Specific Font}
\begin{figure}[t] 
    \centering
    \includegraphics[width=\linewidth]{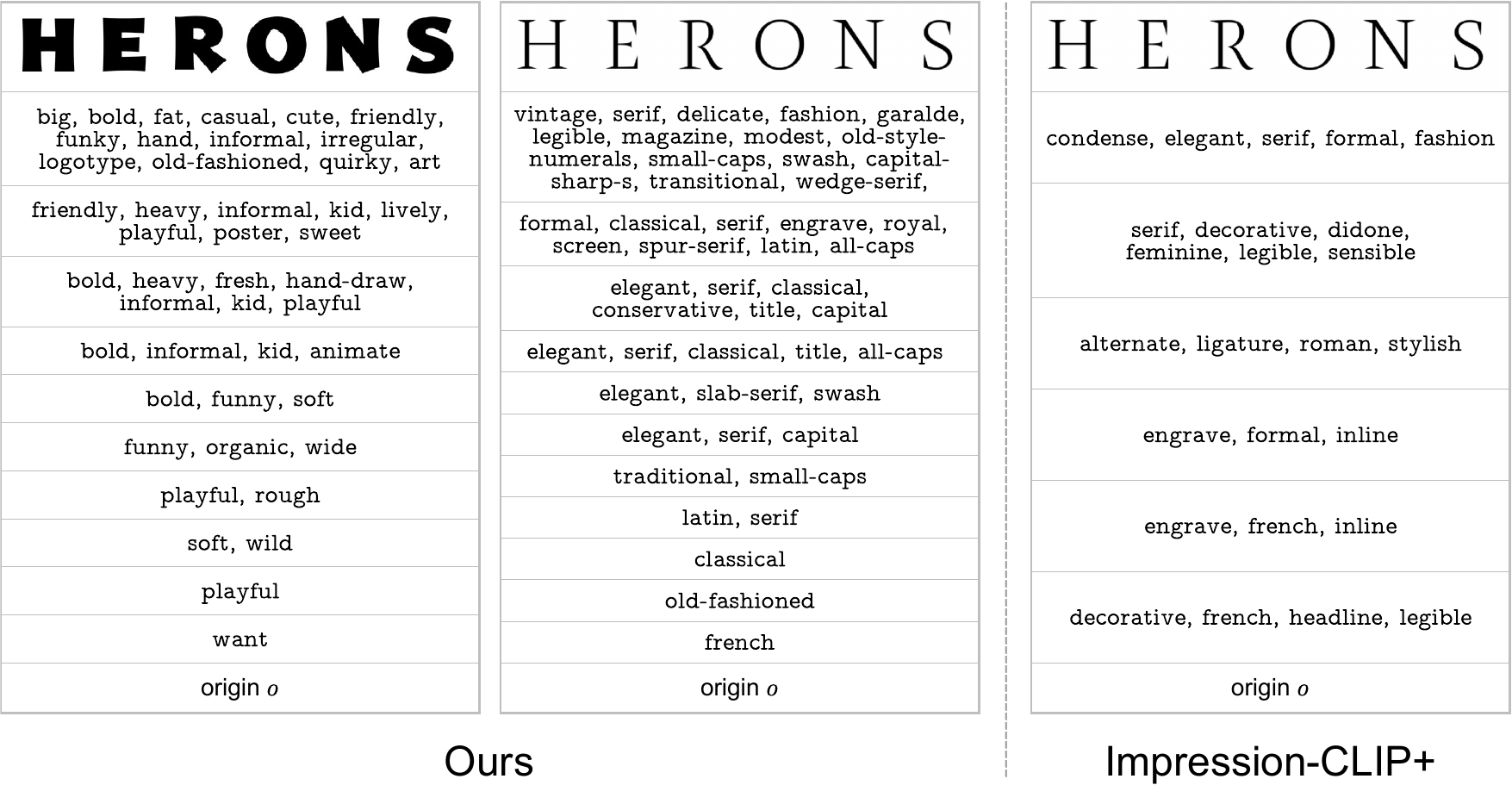}
    \caption{Traversal from the origin $o$ to a target font embedding. Our method shows a coherent shift from abstract to more style-specific impressions while remaining semantically consistent with the target font style.}
    \label{fig:traverse}
\end{figure} 

If the shared latent space properly captures the intended entailment-driven structure between fonts and impressions, then traversing the neighborhood of the trajectory from the origin $o$ to a given font embedding should retrieve impression descriptions with smoothly increasing granularity, from ambiguous to more style-specific. We refer to this procedure as \emph{traversal}. Concretely, we uniformly sample 50 points along the geodesic segment connecting $o$ and the target font embedding, and use each intermediate point as a query to retrieve nearby impression tag sets in the latent space. As retrieval candidates, we construct approximately 500k unique tag sets by repeatedly applying random tag masking to each impression tag set in the dataset. This protocol provides a qualitative test of the intended ordering: low style-specificity impressions should appear near $o$ and become progressively more style-specific as the query approaches the font embedding.
\par

\Cref{fig:traverse} shows traversal results for our method and Impression-CLIP+. For our method, as the query moves from the origin $o$ toward the target font embedding, the retrieved descriptions contain progressively more tags (i.e., become more specific), indicating a clear shift in granularity from abstract to concrete. Moreover, the retrieved descriptions remain semantically consistent with the target font style. In the left example, the results include style-consistent tags such as \IT{bold}, \IT{heavy}, and \IT{fat}, while tags like \IT{playful}, \IT{funny}, \IT{informal}, and \IT{casual} appear consistently along the path from near $o$ to the font. A similar trend is observed in the right example, where tags such as \IT{serif} are retrieved together with increasingly specific impressions containing \IT{classical} and \IT{vintage}.
\par

In contrast, performing the same traversal in Impression-CLIP+ retrieves semantically related tags such as \IT{formal} and \IT{elegant}, but the granularity of the retrieved descriptions remains relatively uniform along the path.
Consequently, the expected progression from ambiguous to more style-specific descriptions as the query moves from the origin $o$ toward the font is not clearly observed.
Overall, these results qualitatively suggest that our method encodes entailment relations between fonts and impressions, as well as an entailment-driven ordering among impressions, while preserving semantic consistency throughout the traversal.
In other words, the learned space is useful for quantifying impression style specificity, namely, how strongly an impression constrains the set of compatible font styles.
\par

\subsection{Style Specificity Analysis of Impression Tags}
\begin{figure}[t] 
    \centering
    \includegraphics[width=\linewidth]{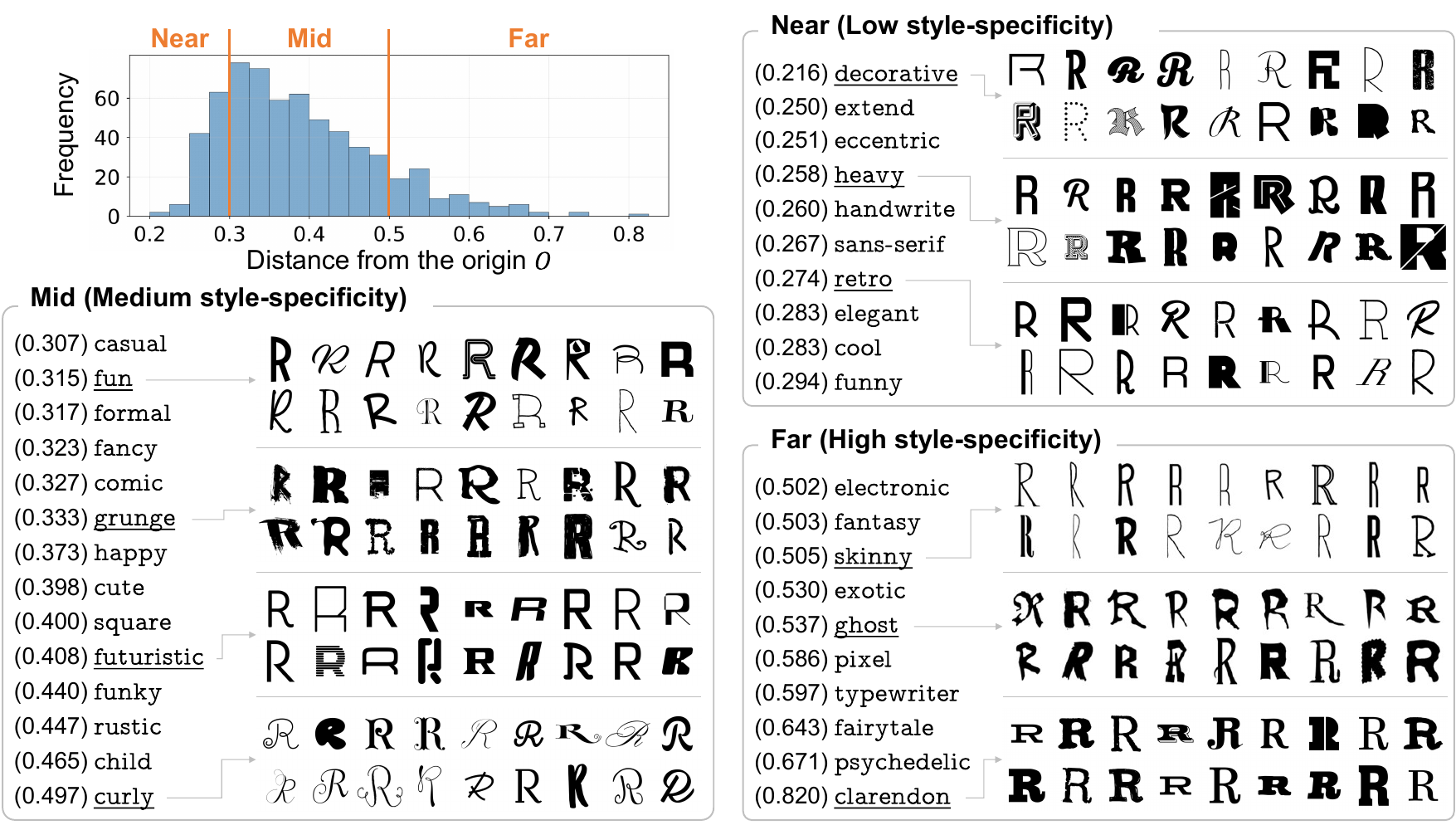}
    \caption{Style specificity of impression tags. The top-left plot shows the distribution of distances from the origin $o$ for tag embeddings, and the remaining panels show example tags with representative fonts. Larger distances indicate higher style specificity. Values in parentheses denote the distance from $o$ (i.e., tag style specificity).}
    \label{fig:single_tag_dist}
\end{figure} 


As shown in \cref{fig:single_tag_dist}, we assess the style specificity of individual impression tags by leveraging a key property of our learned space: distance from the origin $o$ serves as a proxy for specificity. The top-left histogram plots the distance distribution of single-tag embeddings. The results suggest that style specificity is not determined solely by tag count; it also varies substantially across tag types. In particular, some tags lie near $o$, whereas others are embedded much farther away. This spread indicates that impression tags range from ambiguous descriptors compatible with diverse font styles to highly style-specific ones that strongly constrain the set of plausible fonts.
\par

Based on this observation, we analyze the style specificity of each impression tag, namely, how strongly it constrains compatible font styles.
To simplify the analysis, we partition tags into three groups, Near, Mid, and Far, according to their distance from the origin $o$.
We then select representative tags from each group and qualitatively examine their associated fonts to assess how tag specificity relates to font style.
\par

We summarize our main findings below:
\begin{itemize}
    \item The most ambiguous tag is \IT{decorative}, whereas the most style-specific tag is \IT{clarendon}.\footnote{\IT{clarendon} refers to a style of slab serif typefaces originally developed in mid-19th-century Britain for display and advertising purposes.} 
    The associated fonts indicate that \IT{decorative} spans diverse styles, while \IT{clarendon} consistently corresponds to bold slab-serif fonts, supporting the usefulness of our space for assessing tag specificity.

    \item Comparing groups shows that stylistic diversity decreases as the distance from $o$ increases, indicating higher specificity.
    Tags in the Mid group exhibit intermediate stylistic consistency, between the broad variability of Near tags and the strong concentration observed for Far tags.

    \item Near the origin, we often observe adjective-like tags such as \IT{retro} and \IT{elegant}. In contrast, the Far group contains shape-oriented tags such as \IT{skinny} and \IT{typewriter}.

    \item This tendency is not absolute: the Near group also contains tags such as \IT{extend} and \IT{heavy} that intuitively describe geometric properties, yet their associated font styles remain diverse.

    \item Conversely, the Far group includes tags such as \IT{ghost} and \IT{psychedelic}, which are not obviously geometric.
    This suggests that a tag's apparent semantic abstractness does not necessarily determine how strongly it constrains font shapes.
\end{itemize}

Overall, these analyses show that impression tags differ substantially in how strongly they constrain font style: some tags are compatible with a wide range of fonts, whereas others consistently correspond to a narrow and coherent style.
Crucially, this style specificity is not dictated by a tag's surface semantics; instead, it is revealed by the data through the tag's position in the learned space, where distance from the origin $o$ provides a direct and interpretable measure.
This perspective allows us to evaluate font--impression correspondence in terms of constraint strength on the font distribution, rather than relying on semantic intuition alone.
Moreover, the emergence of this clear ordering supports our design choice of an entailment-driven, hierarchy-aware embedding as a suitable framework for jointly modeling fonts and impressions.

\section{Conclusion and Future Works}
We presented a hyperbolic co-embedding framework that models font--impression correspondence through entailment rather than exact one-to-one alignment. Our central observation is that impression descriptions differ in how strongly they constrain font style, and this difference can be represented as \emph{style specificity} in a shared hyperbolic space. To capture this structure, we introduced two complementary entailment relations: impression-to-font entailment and low-to-high style-specificity entailment among impressions. Combined with bidirectional contrastive learning, these constraints yield a shared space in which corresponding fonts and impressions are aligned, while radial distance from the origin provides an interpretable measure of specificity. Experiments on the MyFonts dataset showed improved bidirectional retrieval over baselines, together with clear radial ordering and semantically coherent traversal behavior.

Future work includes extending the learned representation to downstream applications such as font generation and recommendation, incorporating open-vocabulary or natural-language impression descriptions, and validating the estimated style specificity against human judgments. It would also be valuable to examine whether the same entailment-driven structure generalizes to broader typographic settings, including multilingual and non-Latin scripts.

\bibliographystyle{splncs04}
\bibliography{mybibliography}

\end{document}